\documentclass[12pt]{l4dc2021_arxiv} 

\def\eqref#1{equation~\ref{#1}}

\def\1{\bm{1}}

\DeclareMathAlphabet{\mathsfit}{\encodingdefault}{\sfdefault}{m}{sl}
\SetMathAlphabet{\mathsfit}{bold}{\encodingdefault}{\sfdefault}{bx}{n}

\newcommand{\E}{\mathbb{E}}

\usepackage{hyperref}
\usepackage{url}
\usepackage{wrapfig}
\usepackage{stackengine}
\usepackage{mathtools}
\usepackage{wrapfig}
\usepackage{booktabs}

\usepackage{algorithm}

\usepackage{hyperref}
\hypersetup{
 colorlinks=True,
 linkcolor=blue,
 citecolor=blue,
 urlcolor=blue}

\newcommand{\hatT}{\widehat{T}}

\newcommand{\Mstate}{M_\mathcal{S}}
\newcommand{\hatMstate}{\widehat{M}_\mathcal{S}}
\newcommand{\tildeMstate}{\widetilde{M}_\mathcal{S}}

\makeatletter
 \let\Ginclude@graphics\@org@Ginclude@graphics 
\makeatother

\def\shownotes{1}  \ifnum\shownotes=1
\newcommand{\authnote}[2]{{[#1: #2]}}
\else
\newcommand{\authnote}[2]{}
\fi

\newcommand{\Ei}{\bar{\mathbb{E}}}
\newcommand{\Eisub}[1]{\underset{#1}{\Ei}}

\title[Offline RL from Images with Latent Space Models]{Offline Reinforcement Learning from Images\\ with Latent Space Models}
\usepackage{times}

\author{%
 \Name{Rafael Rafailov}\thanks{denotes equal contribution.}~$^{1}$ \Email{rafailov@stanford.edu}\\
 \Name{Tianhe Yu}\footnotemark[1]~$^{1}$ \Email{tianheyu@cs.stanford.edu}\\
 \Name{Aravind Rajeswaran}~$^2$ \Email{aravraj@cs.washington.edu}\\
 \Name{Chelsea Finn}~$^1$ \Email{cbfinn@cs.stanford.edu}\\
 \addr $^1$Department of Computer Science, Stanford University\\
 \addr $^2$School of Computer Science \& Engineering, University of Washington%
}

\begin{document}

\maketitle

\begin{abstract}
Offline reinforcement learning (RL) refers to the problem of learning policies from a static dataset of environment interactions. Offline RL enables extensive use and re-use of historical datasets, while also alleviating safety concerns associated with online exploration, thereby expanding the real-world applicability of RL. Most prior work in offline RL has focused on tasks with compact state representations. However, the ability to learn directly from rich observation spaces like images is critical for real-world applications such as robotics. In this work, we build on recent advances in model-based algorithms for offline RL, and extend them to high-dimensional visual observation spaces. Model-based offline RL algorithms have achieved state of the art results in state based tasks and have strong theoretical guarantees. However, they rely crucially on the ability to quantify uncertainty in the model predictions, which is particularly challenging with image observations. To overcome this challenge, we propose to learn a latent-state dynamics model, and represent the uncertainty in the latent space. Our approach is both tractable in practice and corresponds to maximizing a lower bound of the ELBO in the unknown POMDP. In experiments on a range of challenging image-based locomotion and manipulation tasks, we find that our algorithm significantly outperforms previous offline model-free RL methods as well as state-of-the-art online visual model-based RL methods. Moreover, we also find that our approach excels on an image-based drawer closing task on a real robot using a pre-existing dataset. All results including videos can be found online at \url{https://sites.google.com/view/lompo/}.
\end{abstract}

\section{Introduction}
\label{sec:intro}
\begin{wrapfigure}{r}{0.55\textwidth} 
\vspace{-20pt}
  \begin{center}
    \includegraphics[width=0.549\textwidth]{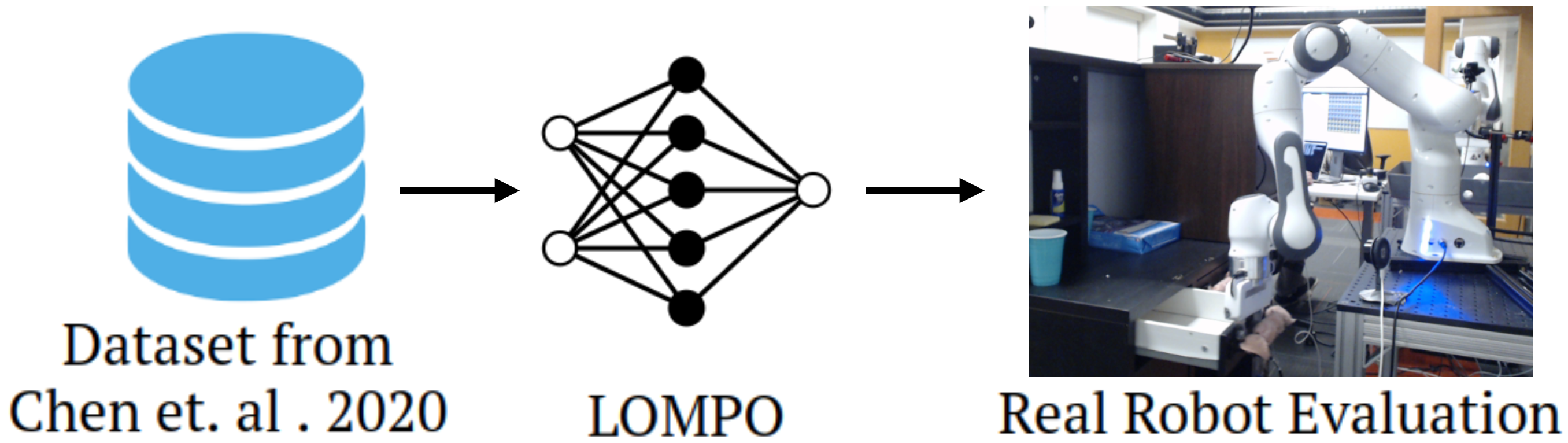}
    \vspace{-0.8cm}
    \caption{\small LOMPO learns vision-based policies from offline datasets, without any interaction in the environment. 
    }
    \label{fig:teaser}
  \end{center}
\vspace{-1.0cm}
\end{wrapfigure} 


For robots and artificial agents to be competent in a wide variety of dynamic and uncertain environments, they require the ability to perceive the world and act based on rich sensory observations like vision. In most real-world scenarios like homes or disaster management, it is difficult to hand-design state representations or simulators, let alone instrument the world to estimate the states. This suggests the need for an end-to-end integration of sensing and control. Despite recent advances~\citep{Dreamer2020Hafner,kostrikov2020image,laskin2020reinforcement}, the interactive (or online) sample complexity for learning control policies from vision is prohibitively high. Furthermore, interactive reinforcement learning~(RL) with physical systems in the real world is fraught with safety challenges that limit widespread applicability. Our goal in this work is to develop approaches for overcoming these challenges by utilizing offline datasets.

Offline RL~\citep{LangeGR12} involves the learning of control policies from static pre-collected data. By using previously collected data, we alleviate the safety challenges associated with online exploration. Large offline datasets are also already available in domains like autonomous driving~\citep{caesar2020nuscenes}, recommendation systems~\citep{harper2015movielens}, and robotic manipulation~\citep{finn2017deep,sharma2018multiple,dasari2019robonet,mandlekar2019scaling},  typically in image (or video) format. Prior work in offline RL (see Section~\ref{sec:related_work}) has typically focused on environments with compact state representations, which are not representative of challenges faced in real-world applications. In this work, we focus on control from pixels using offline datasets. We believe the ability to train vision based policies using offline data can reduce interactive sample complexity, enhance safety, and greatly expand the applicability of RL.

Our work builds on recent advances in model-based offline RL~\citep{kidambi2020morel, yu2020mopo}. 
Model-based RL algorithms have demonstrated impressive sample efficiency in interactive RL~\citep{janner2019trust, RajeswaranGameMBRL, Dreamer2020Hafner}. For offline RL, model-based algorithms~\citep{kidambi2020morel, yu2020mopo} have been shown to be mimimax optimal, obtain state of the art results in a variety of benchmark tasks, as well as generalize to new out-of-distribution tasks. Uncertainty quantification and pessimism have emerged as key principles and requirements for successful offline RL, with both model-based and model-free approaches.
This represents a unique challenge for control from pixels, since directly translating successful approaches in state based domains (e.g. ensembles of models) to image spaces can be computationally prohibitive.


\noindent \paragraph{Our Contributions.}
The main contribution of our work is an algorithm, latent offline model-based policy optimization (LOMPO), which enables learning of visuomotor policies using offline datasets. 
LOMPO can be summarized as follows. (i) Using the available offline data, we learn a variational model with an image encoder, an image decoder, and an ensemble of latent dynamics models. (ii) We construct an uncertainty penalized MDP in the latent state space (which induces a corresponding uncertainty-penalized POMDP in observation space), where we quantify uncertainty based on disagreement between forward models in the latent state space. (iii) We learn a control policy in the learned latent space using the offline dataset by optimizing an uncertainty-penalized objective. The learned uncertainty penalized MDP provides a pessimistic regularizing effect for policy learning and guards against major challenges like distributional shift and model exploitation.
We evaluate our algorithm on four simulated visuomotor control tasks and one real-world robotic manipulation task. We find that LOMPO outperforms or matches prior model-based and model-free methods across the board on these challenging tasks. 

\section{Related Work}
\label{sec:related_work}

Our work is at the intersection of offline RL and control from high-dimensional inputs (i.e. images). We review related work from these fields below.

\noindent \paragraph{Offline RL.} Offline RL has recently emerged as a prominent paradigm for learning control policies~\citep{LangeGR12, levine2020offline}. Most offline RL algorithms augment well known RL algorithms with various forms of regularization. These include regularized variants of importance sampling based algorithms~\citep{LiuSAB19, SwaminathanJ15, nachum2019algaedice, zhanggendice}, actor-critic algorithms~\citep{wu2019behavior, jaques2019way, siegel2020keep, peng2019advantage}, approximate dynamic programming algorithms~\citep{fujimoto2018off, kumar2019stabilizing, kumar2020conservative, Liu2020ProvablyGB, agarwal2019striving}, and model-based RL algorithms~\citep{kidambi2020morel, yu2020mopo, argenson2020model, matsushima2020deployment, swazinna2020overcoming}.
However, most of these prior works focus on problems with low-dimensional compact state information, except that a few that learn to play Atari games~\citep{fujimoto2018off, agarwal2019striving,kumar2020conservative} or propose benchmarks that include simulated locomotion tasks with pixel inputs~\citep{gulcehre2020rl}; our focus in this work is on continuous robot control from high-dimensional perceptual inputs in both simulation and the real world.

\noindent \paragraph{Control from Pixels.}
Control from high-dimensional observation inputs has become an important problem within control and robotics as it makes real world applications more practical. Prior works have tackled this problem with model-free RL methods by learning policies either from pixel inputs end-to-end~\citep{haarnoja2018soft,gelada2019deepmdp, singh2019end, kostrikov2020image, laskin2020reinforcement, VRM2020Han} or on top of unsupervised visual representations~\cite{lange2010deep,ghadirzadeh2017deep,nair2018visual}. Alternatively \citep{SLAC2020Lee, Merlin2018Wayne} train a variational latent space model, but use it only as a filter and train a separate policy on top of the learned latent representation. Model-based RL learns a dynamics model either in the pixel space~\citep{finn2017deep, ebert2018visual} or in a latent space~\citep{levine2016end, finn2016deep, watter2015embed, banijamali2018robust, zhang2019solar, Hafner2019PlanNet, Dreamer2020Hafner, ha2018world, kipf2019contrastive, suraj2020bee} and can either learn a policy within the model or deploy shooting-based planning methods. However, most of those prior works rely critically on online data collection to be successful. Visual foresight algorithms~\citep{finn2017deep,ebert2018visual,suh2020surprising,yen2019experience,suraj2020bee} handle control from pixels in a fully offline setting, but do not explicitly tackle the distributional shift issue that arises; meanwhile, our method is designed to specifically address this. As a result, we find in Section~\ref{sec:real} that our approach significantly outperforms visual foresight.


\newcommand{\D}{{\mathcal{D}}}
\newcommand{\Denv}{{\mathcal{D}_\text{env}}}
\newcommand{\Dmodel}{{\mathcal{D}_\text{model}}}
\newcommand{\given}{\,|\,}
\newcommand{\pib}{\pi^{\textup{B}}}
\newcommand{\Op}{\mathcal{O}}

\section{Preliminaries}
\label{sec:prelim}

\noindent \textbf{POMDPs.} We consider a class of partially observable Markov decision processes (POMDPs), whose transition dynamics can be described using a compact MDP, and observations using an emmision model. Concretely, we consider POMDPs of the form  $M = (\mathcal{X}, \mathcal{S}, \mathcal{A}, T, D, r, \mu_0, \gamma)$ where $\mathcal{X}$ denotes the the visual observation space, $\mathcal{S}$ the unobserved state space, $\mathcal{A}$ the action space, $T(s'|s, a)$ the latent transition distribution, $D(x|s)$ the observation model, $r(s, a)$ the reward function, $\mu_0(s_0)$ the initial latent state distribution, and $\gamma \in (0, 1)$ the discount factor. The goal is to learn a policy $\pi(a_t|x_t)$ that maximizes the discounted expected return $\eta_M \coloneqq \mathbb{E}_\pi[\sum_{t=0}^\infty \gamma^tr(s_t, a_t)]$.

\noindent \paragraph{Control as inference.} For MDPs with directly observable states $s_t$, actions $a_t$, rewards $r_t$, initial state distribution $\mu_0(s_1)$, and the stochastic transition dynamics  $T(s_{t+1}|s_t, a_t)$, the control problem is equivalent to an inference problem in the graphical model with a binary random variable $\Op_t$, which indicates if the agent is optimal at time step $t$. When $p(\Op_t = 1 | s_t, a_t) = \exp(r(s_t, a_t))$, maximizing $p(\Op_{1:H})$ for some finite horizon $H$
via approximate inference is equivalent to optimizing the maximum entropy RL objective, $\E[\sum_{t=1}^H(r(s_t, a_t) + \alpha\mathcal{H}(\pi(\cdot|s_t)))]$~\citep{levine2018reinforcement}. \cite{SLAC2020Lee} extend the framework to POMDPs by factorizing the variational distribution $q(s_{1:H}, a_{t+1:H}|x_{1:t+1}, a_{1:t})$ into a product of inference terms $q(s_{t+1}|x_{t+1}, s_t, a_t)$, latent dynamics terms $T(s_{t+1}|s_t, a_t)$, and policy terms $\pi(a_t|x_{1:t}, a_{1:t-1})$. The evidence lower bound (ELBO) of $p(x_{1:t+1}, \Op_{t+1:H}|a_{1:t})$, the likelihood of observed data and future optimality of the agent is:
\begin{align}
    \log  p(\!~&x_{1:t+1}, \Op_{t+1:H} | a_{1:t})\nonumber \\
    \geq \!~& \E_{(s_{1:H}, a_{t+1:H}) \sim q}
    \left[\sum_{\tau=0}^t\left(\log D(x_{\tau+1}|{s_{\tau+1}}) - D_{KL} \left( q(s_{\tau+1}|x_{\tau+1},s_\tau, a_\tau)\middle\| T(s_{\tau+1}|s_\tau,a_\tau) \right)\right)\right.\nonumber\\
    &\left.+ \sum_{\tau=t+1}^H \left(r(s_\tau, a_\tau) + \log p(a_\tau) - \log\pi(a_\tau|x_{1:\tau}, a_{1:\tau-1})\right)\right]\coloneqq \mathcal{L}_\text{ELBO}\label{eq:policy}
\end{align}
where $r(s_\tau, a_\tau) = \log p(\Op_\tau = 1 | s_\tau, a_\tau)$ and $p(a_\tau)$ is the prior action distribution.

\noindent \paragraph{Offline RL.} In the offline RL problem, the agent must learn a  policy using only a fixed dataset of interactions. In our work, we focus on offline RL in high-dimensional POMDPs, where the agent has access to the fixed dataset $\Denv $ of trajectories, consisting of high-dimensional observations, actions and rewards. The dataset is collected by a  behavior policy $\pib$, which may correspond to a mixture of policies. No additional environment interaction is possible. We call the distribution induced by $\Denv$ as the \textit{behavioral distribution}.

\noindent \paragraph{Model-based offline RL.} 
Model-based RL in conjunction with the key idea of pessimism or conservatism, has emerged as a promising paradigm for offline RL~\citep{kidambi2020morel, yu2020mopo, matsushima2020deployment, argenson2020model}. 
In this work, we build on the MOPO framework~\citep{yu2020mopo}. Given a dataset $\Denv$ from an MDP $M$, MOPO learns a dynamics model $\hatT(\cdot|s,a)$ and reward model $\widehat{r}(s,a)$, and uses these models to construct an uncertainty-penalized MDP $\widetilde{M}$, with dynamics $\hatT$ and a modified reward function $\widetilde{r}(s, a) = \widehat{r}(s, a) - u(s, a)$, where $u(s,a)$ is an estimate of model uncertainty. To account for distribution shift, the policy is optimized in this uncertainty penalized MDP.
With an admissible uncertainty estimator such that $u(s,a) \geq \frac{r_{\max}}{1-\gamma} D_{TV}(T(s,a), \hatT(s,a))$, \citep{yu2020mopo} theoretically show that optimizing a policy under the \textit{uncertainty-penalized} MDP $\widetilde{M} = (\mathcal{S}, \mathcal{A}, \hatT, \widetilde{r}, \mu_0, \gamma)$ is equivalent to optimizing a lower bound of the return under the learned policy in the true MDP $M$. While MOPO achieves impressive results in tasks with low-dimensional observation spaces, it is hard to scale to realistic environments with image observations. In the next section, we present LOMPO, which aims to solve the offline RL problem in a model-based way using high-dimensional observation spaces.


\section{LOMPO: Latent Offline Model-Based Policy Optimization}

Our goal is to design an offline model-based RL method that handles high-dimensional observations. Since we need to learn a model from a fixed dataset without further interaction with the environment, the model predictions will become less trustworthy as the policy rollouts move further from the behavioral distribution. Such inaccurate model predictions would generate observations that could negatively impact the policy optimization (the model exploitation phenomenon). Therefore, quantifying the uncertainty of the observations generated by the learned model is important for offline model-based RL to avoid large extrapolation error on out-of-distribution observations. However, estimating model uncertainty in high-dimensional spaces is challenging: the common approach to uncertainty quantification of learning an ensemble of models is notably memory-intensive and computationally-expensive when applied to visual dynamics models.

In this section, we present our offline visual model-based RL algorithm that address the above challenges by learning a latent dynamics model and estimating the model uncertainty in the compact latent space. Specifically, under the assumption that the control problem in the latent space is an MDP, we construct an uncertainty-penalized latent MDP with the reward penalized by the uncertainty of the latent dynamics model (Section~\ref{sec:quantify}). Next, we construct the corresponding uncertainty-penalized POMDP and optimize the policy and the latent dynamics model in the control as inference framework by maximizing the ELBO in the uncertainty-penalized POMDP, which is a lower bound of the ELBO in the true POMDP (Section~\ref{sec:elbo}). Finally, we discuss our overall practical algorithm LOMPO (Section~\ref{sec:practice}).

\subsection{Quantifying Model Uncertainty in the Latent Space}
\label{sec:quantify}

In order to quantify the uncertainty in the latent state space, we first make the assumption that the latent state space $\mathcal{S}$ forms an MDP, which we define as the latent MDP $\Mstate = (\mathcal{S}, \mathcal{A}, T, \mu_0, \mu_0, \gamma)$ with the same notation from Section~\ref{sec:prelim}. Similarly, we define the estimated latent MDP $\hatMstate = (\mathcal{S}, \mathcal{A}, \hatT, r, \mu_0, \gamma)$ where $\hatT(s'|s,a)$ denotes the learned latent dynamics model. The objective of our algorithm is to learn an optimal policy $\pi(a|s)$ in $\hatMstate$ that also maximizes expected return in $\Mstate$. With the above definitions of $\Mstate$ and $\hatMstate$, we can construct the \textit{uncertainty-penalized} latent MDP $\tildeMstate = (\mathcal{S}, \mathcal{A}, \hatT, \widetilde{r}, \mu_0, \gamma)$ where $\widetilde{r}(s, a) = r(s, a) - \lambda u(s,a)$ where 
$u(s, a)$ is an admissible uncertainty estimator as defined in Section~\ref{sec:prelim}. Under the uncertainty-penalized latent MDP, we first define $\epsilon_u(\pi) = \Eisub{(s,a)\sim \rho^\pi_{\hatT,\mu_0}}[u(s,a)]$ where $\rho^\pi_{T,\mu_0}(s,a) := \sum_{t=0}^\infty \gamma^t \mathbb{P}^\pi_{T, \mu, t}(s)\pi(a \given s)$ denotes the discounted state-action distribution. Then we proceed to show that the total return of a policy $\pi$ under the uncertainty-penalized latent MDP is a lower bound of the the total return under the true latent state MDP and the gap between the learned policy under the uncertainty-penalized latent MDP and the optimal policy $\pi^*$ depends on the latent dynamics model error $\epsilon_u(\pi^*)$. Directly following Theorem 4.4 in \citep{yu2020mopo}, we can show the above claims as following:
\begin{align}
    &\eta_{\Mstate}(\pi) \geq \eta_{\tildeMstate}(\pi),\label{eq:mopo}\\
    &\eta_{\Mstate}(\hat{\pi}) \ge \sup_{\pi}\{\eta_{\Mstate}(\pi) - 2\lambda\epsilon_u(\pi)\}
\end{align}
where $\hat{\pi}$ is the policy learned via maximizing the return under $\tildeMstate$. 
In practice, we do not have access to the uncertainty quantification oracle $u(s, a)$ that upper bounds the latent model error, and we estimate the uncertainty of the latent model using heuristics as discussed in Section~\ref{sec:practice}.

\subsection{Latent Model Training and Policy Optimization with Uncertainty-Penalized ELBO}
\label{sec:elbo}

With the uncertainty-penalized latent MDP defined, we can construct the corresponding uncertainty-penalized POMDP $\widetilde{M} = (\mathcal{X}, \mathcal{S}, \mathcal{A}, \hatT, D, \widetilde{r}, \mu_0, \gamma)$. In the following subsection, we will derive the objectives of the policy learning and latent model learning under the uncertainty-penalized POMDP.

We define the learned variational distribution $\widehat{q}(s_{1:H}, a_{t+1:H}|x_{1:t+1}, a_{1:t})$ as a product of inference terms $q(s_{t+1}|x_{t+1}, s_t, a_t)$, learned latent dynamics terms $\hatT(s_{t+1}|s_t, a_t)$ and policy terms $\pi(a_t|x_{1:t}, a_{1:t-1})$ as follows
\begin{align}
    \widehat{q}(s_{1:H}, a_{t+1:H}|x_{1:t+1}, a_{1:t})\! =\! \prod_{\tau=0}^tq(s_{\tau+1}|x_{\tau+1}, s_\tau, a_\tau)\!\!\!\prod_{\tau=t+1}^{H-1}\!\!\!\hatT(s_{\tau+1}|s_\tau, a_\tau)\!\!\!\prod_{\tau=t+1}^H\!\!\!\pi(a_\tau|x_{1:\tau}, a_{1:\tau-1}).\nonumber
\end{align}
With $\widehat{q}$ and Eq.~\ref{eq:policy}, we can bound the expected return term in Eq.~\ref{eq:policy} from below as follows:
\begin{align}
    &\E_{s_{t+1:H}, a_{t+1:H} \sim q}\left[ \sum_{\tau=t+1}^H r(s_\tau, a_\tau)\right]
    =\Eisub{s_{t+1:H}, a_{t+1:H}\sim \rho^\pi_{T,q(s_{t+1}|x_{t+1}, s_t, a_t)}}\left[ r(s_\tau, a_\tau)\right]\label{eq:init_dist}\\
    &\geq \Eisub{s_{t+1:H}, a_{t+1:H}\sim \rho^\pi_{\hatT,q(s_{t+1}|x_{t+1}, s_t, a_t)}}\left[ \widetilde{r}(s_\tau, a_\tau)\right] = \E_{s_{t+1:H}, a_{t+1:H} \sim \widehat{q}}\left[ \sum_{\tau=t+1}^H \widetilde{r}(s_\tau, a_\tau)\right]\label{eq:apply_mopo}
\end{align}
where Eq.~\ref{eq:init_dist} follows from the definition of $\widehat{q}$ and the discounted state-action distribution with the initial state distribution being  $q(s_{t+1}|x_{t+1}, s_t, a_t)$ and the inequality in Eq.~\ref{eq:apply_mopo} follows from Eq.~\ref{eq:mopo} and the latent MDP assumption. Now we can derive the ELBO in the uncertainty-penalized POMDP from the ELBO in the original POMDP defined in Equation~\ref{eq:policy} as follows:
\begin{align}
    \mathcal{L}_\text{ELBO} 
    &\geq \E_{s_{1:t}, a_{t+1:H} \sim q}\!
    \left[\!\sum_{\tau=0}^t\left(
    \underbrace{\log D(x_{\tau+1}|s_{\tau+1})}_{\mathrm{reconstruction}}\!-\! 
    \underbrace{D_{KL} \left( q(s_{\tau+1}|x_{\tau+1},s_\tau, a_\tau)\middle\| T(s_{\tau+1}|s_\tau,a_\tau)\! \right)}_{\mathrm{consistency}}
    \right)\!\right] \label{eq:model_uncert}\\
    &+ \E_{s_{t+1:H}, a_{t+1:H} \sim \widehat{q}}\left[\sum_{\tau=t+1}^H\left(\widetilde{r}(s_\tau, a_\tau)\! +\! \log p(a_\tau)\! -\! \log\pi(a_\tau|x_{1:\tau}, a_{1:\tau-1})\right)\right] \coloneqq \widetilde{\mathcal{L}}_\text{ELBO}\label{eq:policy_uncert}
\end{align}
\noindent where $\tilde{\mathcal{L}}_\text{ELBO}$ denotes the ELBO in the uncertainty-penalized POMDP, which turns out to be a lower bound of $\mathcal{L}_\text{ELBO}$, the ELBO in the original POMDP. With the uncertainty-penalized ELBO, we optimize the latent dynamics model and the inference model using Eq.~\ref{eq:model_uncert} (the reconstruction and consistency terms), which can be viewed as offline latent model training,
and optimize the policy with Eq.~\ref{eq:policy_uncert}, which uses an uncertainty penalized reward similar to MOPO.
Next, we will discuss the practical implementation of our model training and policy optimization in Section~\ref{sec:practice}.

\subsection{Practical Implementation of LOMPO}
\label{sec:practice}
\begin{figure*}
    \centering
    \includegraphics[width = 0.975\textwidth]{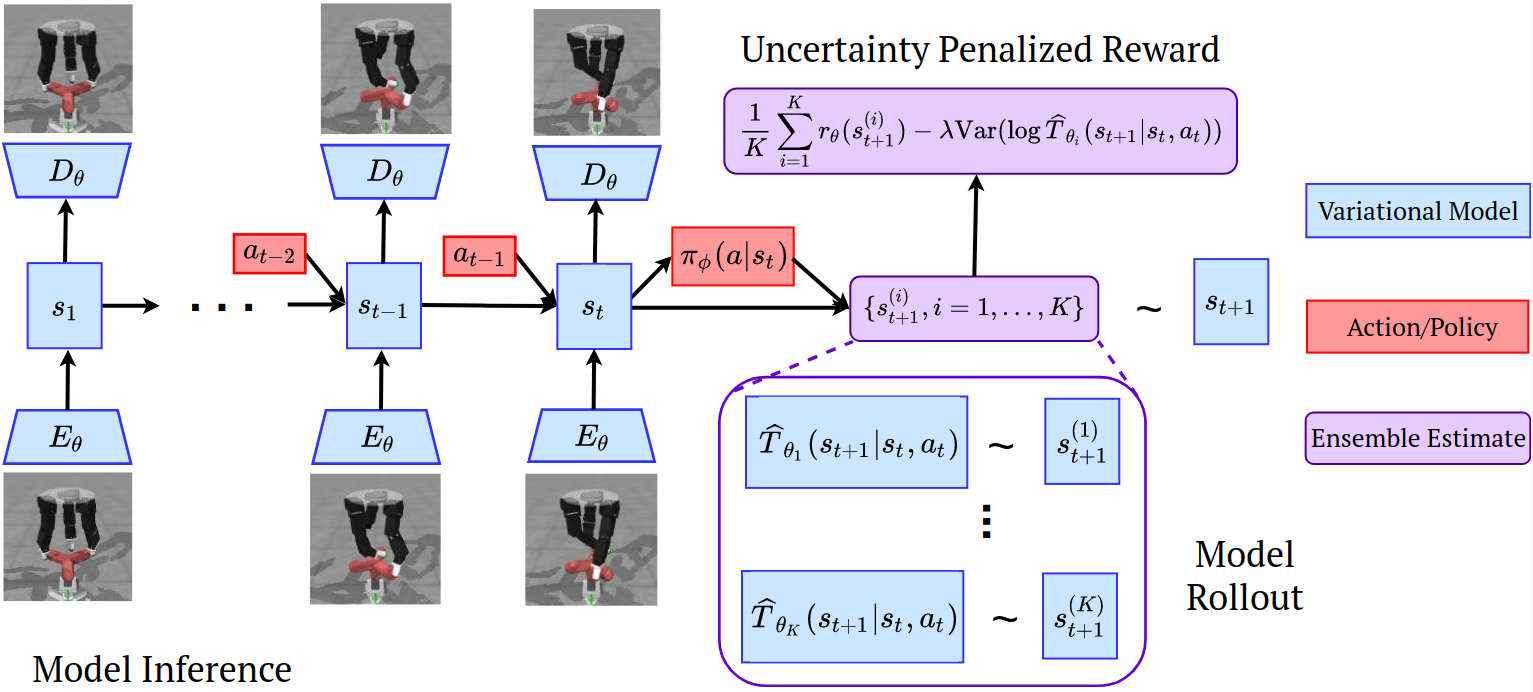}
    \caption{\footnotesize Images are passed through a convolutional encoder $E_\theta$ to form a compact representation which are then used along with previous state to infer the current state $s_t$. The model is trained by reconstructing the images from the latent states through the decoder network $D_{\theta}$. Latent rollouts are carried by choosing a random learned transition model $\hatT_{\theta_j}(s_{t+1}|s_t, a_t)$ and rewards are penalized based on ensemble disagreement.
    }
    \vspace{-0.8cm}
    \label{fig:model}
\end{figure*}



We now present our practical LOMPO algorithm, outlined in Algorithm~\ref{alg:euclid} in Appendix~\ref{app:alg}, and visualize the training framework in Figure~\ref{fig:model}. 

\noindent\textbf{Variational Model Training.}  To estimate the uncertainty term in the uncertainty-penalized POMDP, we train an ensemble of latent transition models and use model-disagreement as a proxy. In designing the optimization approach, we have two main considerations: (1)  we need all of the members of the ensemble to be grounded within the same latent space, (2) we should minimize additional model complexity and training overhead. Considering these, we optimize the following objective:
\begin{equation} \label{eq1}
\begin{split}
 &\sum_{\tau=0}^{H-1}\Big[\mathbb{E}_{q}[\log D(x_{\tau+1}|s_{\tau+1})]-\mathbb{E}_{q}D_{KL}(q(s_{\tau+1}|x_{\tau+1},s_\tau, a_\tau)\| \hatT_{\tau}(s_{\tau+1}|s_{\tau}, a_{\tau}))\Big]
\end{split}
\end{equation}
Here at each step we sample a random learned forward transition model $\hatT_{\tau}$ from a fixed set of $K$ models $\{\hatT_1, \ldots, \hatT_K\}$. The inference distribution $q$ is modeled via the standard mean-field approximation as a uni-modal Gaussian distribution and is shared across all time steps. This explicitly grounds all forward models within the same latent space state representation induced by the inference distribution addressing concern (1) above. Moreover since we use a single forward model at each time step this procedure has the same computational overhead as the regular training of a single variational model, which addresses concern (2).

\noindent \textbf{Model-based Policy Optimization.}
We train a policy and a critic
with components $\pi_{\phi}(a_t|s_t)$ and $Q_{\phi}(s_t, a_t)$ on top of the latent space representation similar to \citep{SLAC2020Lee}, however we deploy model inference for the policy as well as the critic. We made this design choice as it's faster and more efficient to carry out rollouts in the latent space, rather than sampling from the observation model. We maintain two replay buffers $\mathcal{B}_{real}$, $\mathcal{B}_{sample}$.
The real data replay buffer contains transition tuples $s_t, a_t, r_t, s_{t+1}$ from the latent MDP, where states are sampled from the inference distribution $s_{1:H}\sim q(s_{1:H}|x_{1:H}, a_{1:H-1})$ over trajectories from the real dataset $x_{1:H}, r_{1:H}, a_{1:H-1}\sim\mathcal{D}_{env}$.
The latent data buffer contains transitions from rolling-out the policy in the model latent space utilizing the ensemble of learned forward models. During rollouts transitions are carried out at each step by picking a  a random forward model from the ensemble. As discussed in Section~\ref{sec:quantify}, the final rewards for the rollout use ensemble estimates and the uncertainty penalty term and are computed as:
\begin{equation}\label{latentreward}
    \widetilde{r}_t(s_t, a_{t}) = \frac{1}{K}\sum_{i=1}^K r_{\theta}(s_t^{(i)}, a_t) - \lambda u(s_t, a_t)
\end{equation}
where $s^{(i)}_t\sim \hatT_{\theta_i}(s_{t-1}, a_{i-1})$ are sampled from each forward model and $s_t$ is sampled from $\{s^{(i)}_t, i=1,\ldots, K\}$. Here $u(s_t, a_t)$
is an estimate of model uncertainty and $\lambda$ is a penalty parameter. In particular, we pick $u(s_t, a_t)$ as disagreement of the latent model predictions in the ensemble, i.e. the variance of log-likelihoods under the ensemble
$u(s_t, a_t) = \text{Var}(\{\log \hatT_{\theta_i}(s_t|s_{t-1}, a_{t-1}), i=1,\ldots, K\}),$
since ensembles have been shown to capture the epistemic uncertainty~\citep{bickel1981some} and also work well for model-based RL in practice~\citep{yu2020mopo, kidambi2020morel}. We used this heuristic to estimate uncertainty as it estimates disagreement across the means of the forward models, as well as the variances.
Finally the actor and critic $\pi_{\phi}(a_t|s_t), Q_{\phi}(s_t, a_t)$ are trained using standard off-policy training algorithm using batches of equal data mixed from the real and sampled replay buffers, as we find that maintaining a fixed sampling proportion is important in preventing distributional shift in the actor-critic training. 


\section{Experiments}

The goal of our experimentation evaluation is to answer the following questions. (1) Can offline RL reliably scale to realistic robot environments with complex dynamics and interactions? (2) How does LOMPO compare to prior offline model-free RL algorithms and online model-based RL algorithms when learning vision-based control tasks from offline data? (3) How does the quality and size of the dataset affect performance? (4) Can LOMPO be applied to an offline RL task on a real robot with raw camera image observations? We answer questions (1), (2) and (3) in Section~\ref{sec:sim} and address question (4) in Section~\ref{sec:real}. All implementation details such as model architectures and hyperparameter choices are included in Appendix~\ref{app:details}.


\begin{figure}
    \centering
    \includegraphics[width = 0.195\textwidth]{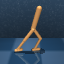}
    \includegraphics[width = 0.195\textwidth]{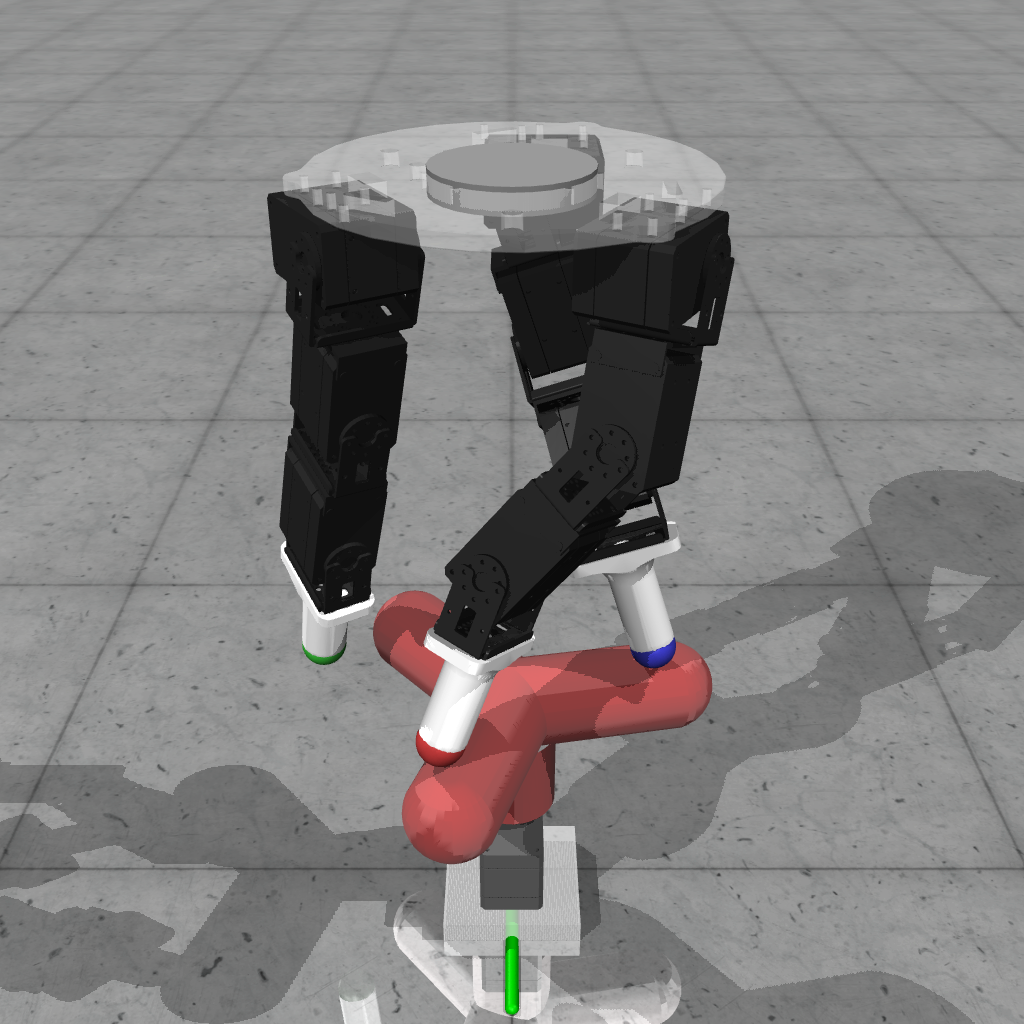}
    \includegraphics[width = 0.195\textwidth]{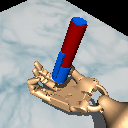}
    \includegraphics[width = 0.195\textwidth]{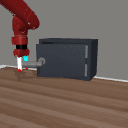}
    \includegraphics[width = 0.195\textwidth]{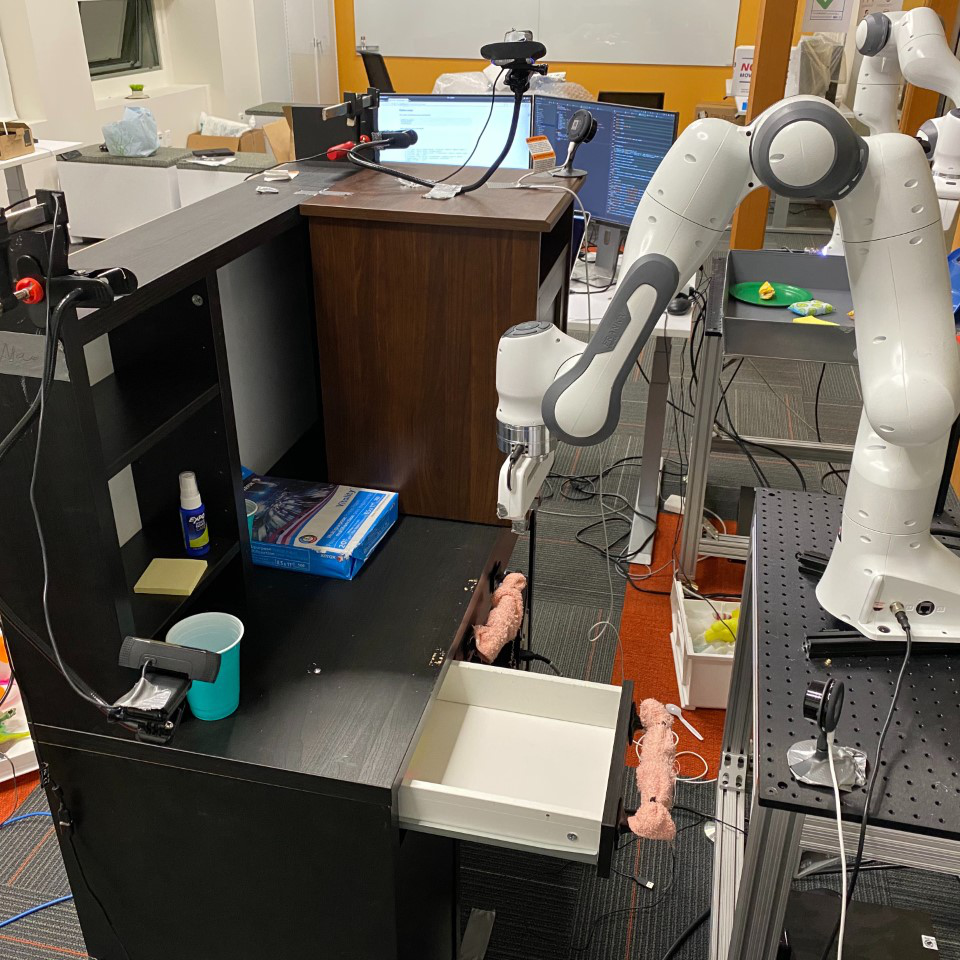}
    \vspace{-0.6cm}
    \caption{\footnotesize Test environments: DeepMind Control Walker task - the observations are raw $64\times64$ images. Robel D'Claw Screw and Adroit Pen tasks observations are raw $128\times128$ images and robot proprioception. Sawyer Door open environment - the observation space is raw $128\times128$ images. The observations for the real robot environment are raw $64\times 64$ images from the overhead camera.}
    \label{fig:envs}
    \vspace{-0.8cm}
\end{figure}

\subsection{Simulated Experiments}
\label{sec:sim}

Previous offline RL benchmarks are not well-suited for answering questions (1), (2) and (3) above as they largely lack image-based robot control problems. Thus, to answer those three questions, we design a suite of four simulated image-based offline RL problems, focusing on robotics applications, described in Appendix \ref{app:envs} and visualized in the four pictures on the left in Figure~\ref{fig:envs}. We also include the visualization of the samples generated from our learned variational model on the four environments in Appendix~\ref{app:latent_samples}. All of the environments and datasets will be open-sourced to allow future work to also study this problem and make direct comparisons.

\noindent \textbf{Comparisons.} We compare our proposed method to both model-free and model-based learning algorithms. Our first comparison is direct behavior cloning (BC) from raw image observations, which has proved to be a strong baseline in the past (\cite{fu2020d4rl}). We also benchmark to Conservative Q-Learning (\cite{kumar2020conservative}), which is a state of the art offline learning algorithm in the low-dimensional case (\cite{fu2020d4rl}); however, we again train it from raw image observations. We evaluate an MBPO based model (\cite{janner2019trust}), which also carries out policy rollouts in latent space similar to LOMPO, but does not apply an uncertainty penalty. The performance of this method is indicative of online model-based methods \citep{Hafner2019PlanNet, Dreamer2020Hafner}. We also train the Stochastic Latent Actor Critic (SLAC) model (\cite{SLAC2020Lee}), a state of the art online learning algorithm from images, however we train fully online. The goal of this benchmark is to evaluate the need for representation learning in offline RL.

\noindent \textbf{Results.} Results are reported in Table \ref{results}. We see that LOMPO achieves high-scores across most high-fidelity simulation environments, using raw observations. Moreover, our proposed model outperforms other model-based learning algorithms across the board and is the only model-based learning algorithm that achieves any success on several environments. Comparing to model-free algorithms, LOMPO still outperforms CQL and behaviour cloning across most environments, with the exception of learning on the expert dataset on the Door Open task. This is a well-known phenomenon when learning dynamics models from narrow expert data. On the other hand, given the thin data distribution and relatively simple dynamics of the task, direct behavior cloning from images performs well on both the medium-expert and expert dataset. We hypothesize that LMOPO performs well on the D'Claw and Adroit expert datasets, as these environments are relatively stationary, as compared to a robot arm manipulation task, and even actions from a stochastic expert cover a wide range of the environment dynamics. The question of dataset size (question (3)) has not been extensively studied in offline RL, however we found that this almost as important as the data distribution itself. We believe this is an important question as collecting large robot dataset can still be complex and expensive task. We carried additional ablation experiments (Appendix \ref{app:ablation}) and discover that performance for regular model-based and model-free methods can decline drastically as data size decrease, while LOMPO performance remains relatively stable.


\begin{table*}[t]
\footnotesize
\centering
\begin{tabular}{p{2.5cm}|p{2.5cm}|c|c|c|c|c}

\toprule
\textbf{Environment} & \textbf{Dataset} & \textbf{LOMPO (ours)} & \textbf{LMBRL} & \textbf{Offline SLAC} & \textbf{CQL} & \textbf{BC}\\
 \midrule
Walker Walk   & medium-replay &\textbf{74.9}&   44.7 & -0.1 & 14.7 & 5.3\\
Walker Walk   & medium-expert &\textbf{91.7}&   76.3 & 32.8 & 45.1 & 15.6\\
Walker Walk   & expert        &\textbf{75.8}&   24.5 & 11.3 & 40.3 & 11.8\\

D'Claw Screw  & medium-replay &\textbf{71.8}& \textbf{72.4}   & 65.9 & 26.3 & 11.7\\
D'Claw Screw  & medium-expert &\textbf{100.4}&   \textbf{96.2}& 76.3 & 30.3 & 27.6\\
D'Claw Screw  & expert        &\textbf{99.2} &  90.8& 63.4 & 24.2& 25.2\\
Adroit Pen  & medium-replay &\textbf{82.8} & 5.2   & 5.4 & 25.8&46.7\\
Adroit Pen  & medium-expert &\textbf{94.6}&   0.0& -1.7 & 43.5&41.8\\
Adroit Pen  & expert        &\textbf{96.1} &  0.2& -0.4 & 51.4&45.4\\

Door Open  & medium-expert &\textbf{95.7}&   0.0& 0.0 & 0.0&72.2\\
Door Open  & expert        &0.0          &   0.0& 0.0 & 0.0&\textbf{97.4}\\

 \bottomrule
\end{tabular}
\caption{\footnotesize Results for the DeepMind Control Walker task, the Robel D'Claw Screw task, Adroit Pen task and the Sawyer Door Open task. The scores are undiscounted
average returns normalized to roughly lie between 0 and 100, where a score of 0 corresponds to a random policy,
and 100 corresponds to an expert. LOMPO consistently outperforms LMBRL, offline SLAC, CQL, and behavioral cloning in almost all settings.}
\vspace{-0.8cm}
\label{results}
\end{table*}

\subsection{Real Robot Experiments}
\label{sec:real}
To answer question (4), we deploy LOMPO on a real Franka Emika Panda robot arm. 

\noindent \textbf{Task.} The environment consists of a Panda arm mounted in front of an Ikea desk cluttered with random distractor objects. The robot arm is initialized randomly above the desk and the drawer is initialized randomly in a open position. The goal of the robot is to navigate to the handle, hook it, and close the drawer. Observations are raw RGB images from a single overhead camera. The complete setup is shown in the rightmost picture in Figure~\ref{fig:envs}.

\noindent \textbf{Dataset.}
We use a pre-existing dataset of 1000 trajectories that was collected using a semi-supervised batch exploration algorithm~\citep{suraj2020bee}. A small balanced dataset of 200 images (0.2\% of the full dataset) is manually labeled with whether the drawer is open or closed.
Following the set up by~\citep{suraj2020bee}, we use this dataset to train a classifier to predict whether the drawer is open or closed. We use the classifier probability as a reward for RL, which leads to a sparse, noisy, and unstable reward signal, which is reflective of one challenge of real-world RL.  
Since the dataset was collected and labeled in the context of a different paper~\citep{suraj2020bee}, this experiment evaluates the ability to reuse existing offline datasets, which further exemplifies real-world problems.

\noindent \textbf{Comparisons.} We compare LOMPO, LMBRL, Offline SLAC, and visual foresight~\citep{finn2017deep} using an SV2P model \citep{babaeizadeh2018stochastic} and a CEM planner. 
As CQL did not achieve competitive performance on the simulated environments, we did not deploy it on the real robot. Moreover the offline dataset has high variance and consists of mostly non-task centric exploration, which is not suitable for imitation; hence we also did not evaluate behavioral cloning.

 \begin{wraptable}{r}{5cm}
\footnotesize
\centering
\vspace{-0.8cm}
\caption{\footnotesize Results for the Franka desk drawer-closing task.}
\vspace{-0.2cm}
\begin{tabular}{ l|c}
\toprule
Method & Success\\
\midrule
LOMPO (ours) & \textbf{76.0\%} \\
LMBRL & 0.0\% \\
Offline SLAC & 0.0\% \\
Visual Foresight & 0.0\% \\
 \bottomrule
\end{tabular}
\vspace{-0.4cm}
\label{real_results}
\end{wraptable}
\noindent \textbf{Results.}  We carry out 25 evaluation rollouts on the real robot\footnote{Evaluation videos are available at \url{https://sites.google.com/view/lompo/}.} and summarize the results in Table~\ref{real_results}. Overall, 24/25 of the LOMPO agent rollouts successfully navigate to the drawer handle, hook it, and push the drawer in; however the agent fully closes the drawer in only 19 of the rollouts for a final success rate of \textbf{76\%}. We hypothesize that the agent does not always close the door as the classifier reward incorrectly predicts the drawer as closed when the drawer is slightly open. In contrast, the LMBRL, Offline SLAC, and visual foresight agents do not manage to successfully navigate to the correct handle location, hence achieving a success rate of \textbf{0\%}. These experiments suggest that LOMPO's uncertainty estimation and pessimism are critical for good offline RL performance. Finally, we note that \citep{suraj2020bee} evaluate visual foresight in the same environment but on an easier version of this task, where the robot arm is initialized near the drawer handle. In this shorter-horizon problem, visual foresight achieves a success rate of \textbf{65\%} (Figure 8 of~\cite{suraj2020bee}), which is still lower than LOMPO's success rate in the more difficult setting. Hence, this suggests that LOMPO is better able to solve problems with a longer time horizons by incorporating pessimism and a learned value function.


\section{Conclusion}

We present an offline model-based RL algorithm that handles high-dimensional observations with latent dynamics models and uncertainty quantification.
We noted that learning a visual dynamics model is challenging and quantifying uncertainty in the pixel space is extremely costly. We address such challenges by learning a latent space and quantify the model uncertainty in the latent space. Our algorithm, LOMPO, penalizes latent states with latent model uncertainty implemented as the latent model ensemble disagreement. LOMPO empirically outperforms previous latent model-based and model-free models in the offline setting on four simulated locomotion and manipulation tasks and one real-world robotic manipulation task. A potential direction for future work is to apply LOMPO to the multi-task setting. Having a single shared model trained with data from multiple tasks can help learn more accurate vision and dynamics models, thus improving performance, sample efficiency, and generalization.

\acks{We want to thank Suraj Nair for sharing the BEE dataset with us and his help with setting up the Panda drawer environment. This work was supported in part by ONR grant N00014-20-1-2675 and Intel Corporation. CF is a CIFAR Fellow in the Learning in Machines and Brains program. Aravind Rajeswaran was supported by a JP Morgan PhD Fellowship (2020).}

\bibliography{references}
\newpage
\appendix

\section{Environments and Datasets}
\label{app:envs}
We describe the simulated environments and the dataset collection in more detail below. 

\noindent \textbf{Environments and Tasks.} We evaluate our method in four simulated environments. 

\begin{enumerate}
    \item  The first task is the standard walker task from the DeepMind Control suite~\citep{tassa2020dmcontrol}. The observations consist of raw $64\times 64$ images, similiar to previous works. We apply an action repeat of 2 over the base environment. This is a standard practice \citep{SLAC2020Lee, Dreamer2020Hafner} as timestep in these environments is relatively short and leaves low visual footprint, which makes model learning hard.
    
    \item The second experiment consists of a modified version of the D'Claw screw task from the Robel benchmark \citep{ahn2019robel}, where the goal of the robot is to continuously turn the valve as fast as possible. The agent receives a dense negative penalty for positioning the fingers and a sparse reward whenever it turns the valve.  The observation space consist of robot proprioception and $128\times 128$ raw images. We apply an action repeat of 2 over the base environment.
    
    \item The third environment is based on the Adroit pen task \citep{Rajeswaran-RSS-18} with a fixed goal, which requires the agent to flip the pen around and catch it at  certain angle. The observation space  consist of robot proprioception and $128\times 128$ raw images. We apply an action repeat of 4 over the base environment.
    
    \item The final simulation experiment is based on a Sawyer manipulation task~\citep{yu2020meta}, which requires opening a door. The agent received a sparse reward when the door is fully opened and no reward otherwise. The goal of this environment is to test learning in a realistic multi-stage robot arm environment with a sparse reward. This is a hard environment, not solvable with online RL. The observation space consists of $128\times 128$ images without access to the robot state. We apply an action repeat of 4 over the base environment.
    
\end{enumerate}
    
We provide visualizations of all simulated tasks in Figure~\ref{fig:envs}. Each of these environments presents different challenges. In the Walker task the agent needs to learn a forward dynamics model completely from images, which are generated by a non-stationary camera. The D'Claw and Adroit environments are both quite challenging as the model needs to merge proprioception and visual information in order to estimate a hard contact model with realistic physics, as well as forward dynamics under a high-dimensional action space (9 and 26 respectively) and a sparse reward function (for the calw environment). The Sawyer arm environment requires learning a 3D dynamics model without access to depth estimation and a sparse reward function, both of which are common in real world RL.


\noindent\textbf{Datasets.}   We construct new sets of offline datasets with image observations. Similar to the protocol by \citet{fu2020d4rl}, we create three types of datasets for each task, which are obtained by training an agent using soft-actor critic~\citep{haarnoja2018soft} from the ground-truth state and recording the corresponding image observations, actions, and rewards. 
The \emph{medium-replay} datasets consist of data from the training replay buffer up to the point where the policy reaches performance of about half the expert level performance. The goal of these datasets is to test learning on incomplete training data. 
The \emph{medium-expert} datasets consist of the second half of the replay buffer after the agent reaches medium-level performance. The goal of these datasets is to test learning on a mixture of data from sub-optimal policies. 
The \emph{expert} dataset consist of data sampled from the stochastic SAC expert policy. The goal of these datasets is to test learning on a thin data distribution. The dataset sizes for Walker, D'Claw, Adroit and the Sawyer environments are 100K, 180K, 1M and 50K transitions respectively.

\section{Ablation Studies on Varying Offline Dataset Size}
\label{app:ablation}

We test the effect of dataset size, given that the data is sampled from the same distribution. We create a medium-expert dataset of size 1M on the D'Claw Screw environment by mixing data from 3 separate policy training runs. We then created two more datasets by sub-sampling by a factor of 5 and 25 respectively. We observe that LMOPO still performs well in the low-data regime, as compared to regular latent model-based RL and Offline SLAC. 

\begin{figure}[H]
    \centering
    \includegraphics[width = 0.75\textwidth]{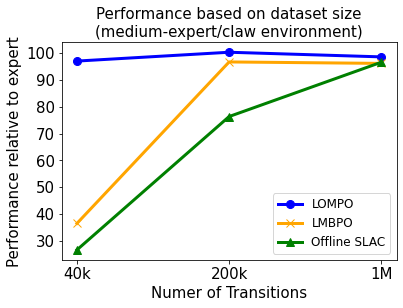}
    \caption{Agent performance based on dataset size}
    \label{fig:ablation}
    \vspace{-0.5cm}
\end{figure}

\section{Implementation details}
\label{app:details}

The latent dynamics model and the observation model consist of the following components as in \citep{Dreamer2020Hafner}:
\begin{gather}
\begin{aligned}
&\text{Image encoder:} &&h_t=E_\theta(x_t) \\
&\text{Inference model:} && s_t \sim q_\theta(s_t|h_t, s_{t-1}, a_{t-1})\\
&\text{Latent transition model:} && s_t \sim \hatT_\theta(s_t| s_{t-1}, a_{t-1})\\
&\text{Reward predictor:} && r_t \sim p_\theta(r_t|s_t) \\
&\text{Image decoder:} && x_t \sim D_\theta(x_t|s_t).
\label{eq:latent_model}
\end{aligned}
\end{gather}%
where $x_t \in \mathcal{X}, s_t \in \mathcal{S}, a_t \in \mathcal{A}$ are the environment observation, latent state, action at time $t$ respectively.
We use $\theta$ to denote the concatenation of all the parameters involved in the latent space dynamics model. The latent dynamics model is represented by a RSSM (\cite{Hafner2019PlanNet}). Specifically, we adopt the latent space representation $s_t = [d_t, z_t]$, which consists of a deterministic $d_t$ and a sampled stochastic representation $z_t$. With such a latent space representation, we use the following components:
\begin{gather}
\begin{aligned}
&\text{Deterministic State Model:} &&d_t = f_{\theta}(d_{t-1}, z_{t-1}, a_{t-1}) \\
&\text{Stochastic Inference Model:} &&z_t \sim q_\theta(z_t|h_t, d_t)\\
&\text{Ensemble of Transition Models:} &&z_t \sim p_{\theta_k}(z_t| d_t, z_{t-1}, a_{t-1})\\
\label{eq:rssm}
\end{aligned}
\end{gather}%
 where $h_t = E_\theta(x_t)$ are observation features as defined in Eq.~\ref{eq:latent_model}. The deterministic representation $f_{\theta}$ is implemented as a single GRU cell and is shared between the forward and inference models. All the learned forward models $\hatT_{\theta_k}, k=1,\ldots,K$ in the ensemble share the same deterministic model $f_{\theta}$ but separate stochastic transition models  $p_{\theta_k}, k=1,\ldots,K$, which are implemented as MLPs. Finally the stochastic inference model is also implemented as an MLP network.

The encoder network $E_{\theta}$ is modeled as a convolutional neural network. For the DeepMind Control Walker task the network has 4 layers with [32, 64, 128, 256] channels respectively. For the D'Claw Screw, Adroit Pen and Door Open environments the convolutional model has 5 layers with [32, 64, 128, 256, 256] channels. All kernels have size 4 and stride 2. The reconstruction model $D_{\theta}$ for DeepMind Control walker task has 4 layers with [128, 64, 32, 3] channels with kernel size [5, 5, 6, 6] and stride 2. For the other environments the reconstruction network has 5 layers with [128, 64, 32, 32, 3] channels with kernel size [5,5,5,4,4] and stride 2. The reward reconstruction network is a two-layer fully connected network. The deterministic path $f_{\theta}$ of the RSSM is modeled as a GRU cell with 256 units for all environments, except the Adroit Pen task, which uses 512 units. All the forward models $T_{\theta_i}, i=1,\ldots, K$ and inference $q_{\theta}$ models are 3-layer fully-connected networks with 256 units, except for the Adroit Pen task, which uses 512. The variational model is trained with the Adam optimizer with $lr = 6e-4$.

Both the actor and the critic are modeled as fully-connected networks with 3 layers and 256 units. We use the Adam optimizer with $lr = 3e-4$.

\section{Main Algorithm}
\label{app:alg}

We present the full LOMPO algorithm in Algorithm~\ref{alg:euclid}.

\begin{algorithm}[h]
\DontPrintSemicolon
\caption{LOMPO}\label{alg:euclid}
\KwIn{model train steps, initial real steps, initial latent steps, number of epochs, epoch real batch, epoch latent batch, number of models $K$}
\For{model train steps}{
 Sample a batch of sequences of raw observations $(o_{1:H}, a_{1:H-1}, r_{1:H-1})$ from batch dataset $\Denv$ and train variational ensemble model using equation \ref{eq1} with $K$ models\;
}

\While{size $\mathcal{B}_{real}<$ initial real steps}{
Sample a batch of sequences $(o_{1:H}, a_{1:H-1}, r_{1:H-1})$ from batch dataset $D_{env}$\; 

Sample latent states $s_{1:H}\sim q_{\theta}(s_{1:H}|o_{1:H}, a_{1:H-1})$ from the trained inference model\;

Add batches $s_t, a_t, s_{t+1}, r_{t}$ to real replay buffer $\mathcal{B}_{real}$\;
}

\While{size $\mathcal{B}_{latent}<$ initial latent steps}{
Sample a batch of sequences $(o_{1:H}, a_{1:H-1}, r_{1:H-1})$ from batch dataset $\Denv$\;

 Sample a set of latent states $\mathbf{S} \sim q_{\theta}(s_{1:H}|o_{1:H}, a_{1:H-1})$ from the trained inference model\;
    \For{$s_0 \in \mathbf{S}$} {
        \For{$h \in \{1:H\}$}{
            Sample a latent transition model from $p_{\theta_j}$ from  $i=1,\ldots, K$\;
            
            Sample a random action $a_{h-1}$ and next state $s_h\sim \hatT_{\theta_j}(s|s_{h-1}, a_{h-1})$\;
            
            Compute reward $\tilde{r}_{h-1}$ using equation \ref{latentreward} and add $(s_{h-1}, a_{h-1}, s_h, \tilde{r}_{h-1})$ to $\mathcal{B}_{latent}$\;
        }
    }
}

\For{number of epochs}{
\For{number actor-critic steps}{
 Equally sample batch $(\mathbf{s_t}, \mathbf{a_t}, \mathbf{r_t}, \mathbf{s_{t+1}})$ from $\mathcal{B}_{real}\cup \mathcal{B}_{latent}$\;
 
 Update $Q_{\phi}(s, a), \pi_{\phi}(a|s)$ using any off-policy algorithm\;
}

\For{epoch real batch}{
 Sample a batch of sequences $(o_{1:H}, a_{1:H-1}, r_{1:H-1})$ from batch dataset $D_{env}$\;
 
 Sample latent states $s_{1:H}\sim q_{\theta}(s_{1:H}|o_{1:H}, a_{1:H-1})$ from the trained inference model\;
 
 Add batches $s_t, a_t, s_{t+1}, r_{t}$ to real replay buffer $\mathcal{B}_{real}$\;
}

\For{epoch latent batch}{
Sample a batch of sequences $(o_{1:H}, a_{1:H-1}, r_{1:H-1})$ from batch dataset $\mathcal{D}$\;

 Sample a set of latent states $\mathbf{S} \sim q_{\theta}(s_{1:H}|o_{1:H}, a_{1:H-1})$ from the trained inference model\;
    \For{$s_0 \in \mathbf{S}$}{
        \For{$h \in \{1:H\}$}{
             Sample a latent transition model from $\hatT_{\theta_j}$ from  $i=1,\ldots, K$\;
             
             Sample an action $a_{h-1} \sim \pi_{\phi}(a|s_{h-1})$ and next state $s_h\sim \hatT_{\theta_j}(s|s_{h-1}, a_{h-1})$\;
             
             Compute reward $\tilde{r}_{h-1}$ using equation \ref{latentreward} and add transition to $\mathcal{B}_{latent}$\;
        }
    }
}
}
\end{algorithm}

\section{Variational Latent Model Samples}
\label{app:latent_samples}
\begin{figure}[ht]
    \centering
    \includegraphics[width = 0.975\textwidth]{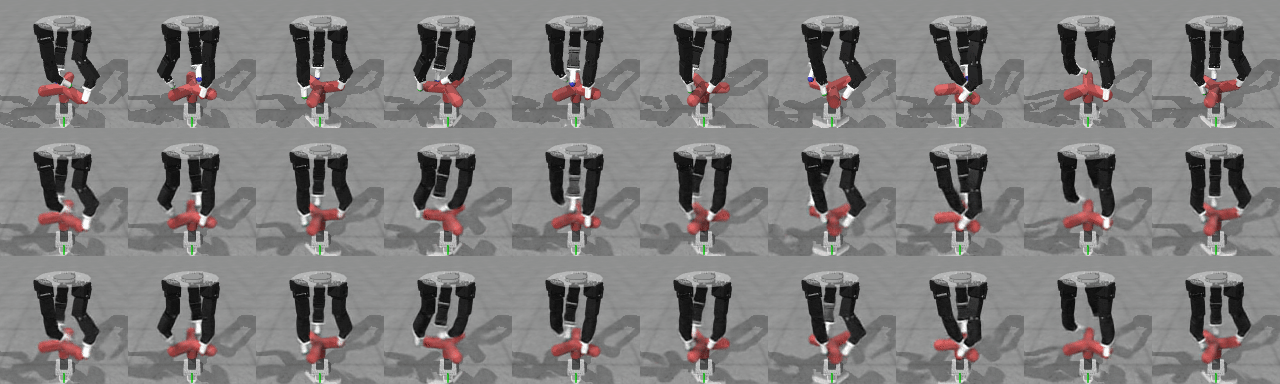}
    \includegraphics[width = 0.975\textwidth]{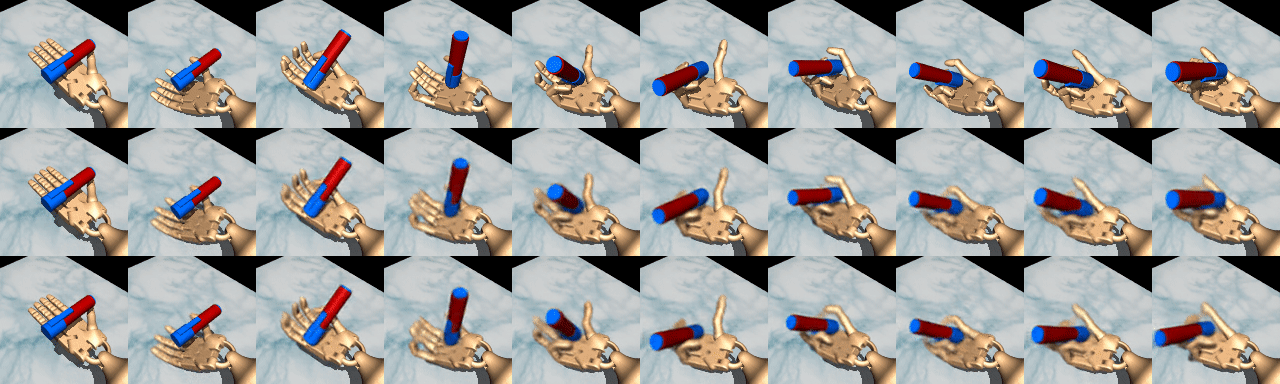}
    \includegraphics[width = 0.975\textwidth]{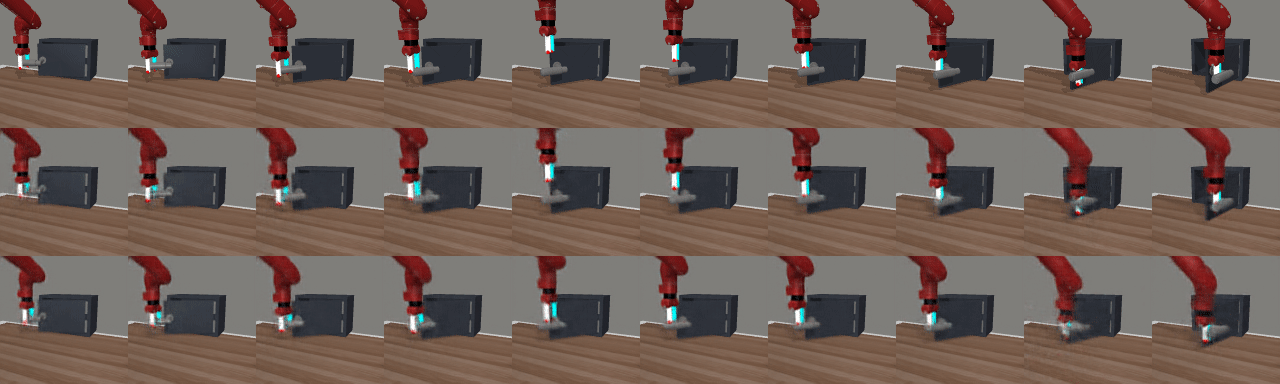}
    \includegraphics[width = 0.975\textwidth]{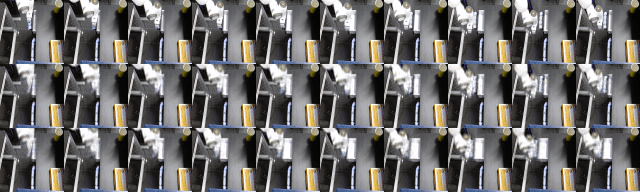}
    \caption{Samples from the learned variational model. Fist row: ground truth sequence; second row: posterior model samples; third row: ensemble latent model rollout conditioned on the action sequence.}
    \label{fig:samples}
\end{figure}

For samples generated by our variational latent models, see Figure~\ref{fig:samples}.

\end{document}